\documentclass[11pt]{article}
\usepackage[margin=1in]{geometry}
\usepackage{amsmath,amssymb}
\usepackage{graphicx}
\usepackage{booktabs}
\usepackage{array}
\usepackage[colorlinks=true,linkcolor=blue,citecolor=blue,urlcolor=blue]{hyperref}
\usepackage{natbib}
\usepackage{caption}
\title{Beyond Coordinate Gauge: An Audited Protocol for Detecting Donor-Specific Functional Fingerprints after Neural Collapse}
\author{
Truong Xuan Khanh\textsuperscript{1}\quad Phan Thanh Duc\textsuperscript{2}\\[4pt]
\textsuperscript{1}H\&K Research Studio, Clevix LLC, Hanoi, Vietnam\\
\textsuperscript{2}Banking Academy of Vietnam, Hanoi, Vietnam\\[2pt]
\texttt{khanh@clevix.vn}\quad \texttt{ducpt@bav.edu.vn}
}
\date{}

\begin{document}
\maketitle

\begin{abstract}
Independently trained neural networks have, by construction, no shared
neuron-index reference frame, so direct cross-trajectory comparison
requires accounting for coordinate freedom. Neural Collapse sharpens
rather than resolves this problem: independently trained networks
converge toward a shared, low-dimensional geometry, raising the question
of whether trajectory-specific functional variation remains
distinguishable after that convergence. We separate three empirical
claims -- detectability, transplantability, and causal persistence -- and
address the first. Using five independently trained MLP-5 networks in a
paper-matched reconstruction of Neural Collapse on MNIST, we fit an
orthogonal Procrustes alignment and apply the corresponding affine
correction to map donor classifier heads into recipient coordinates. The
affine head transformation preserves donor logits exactly under the
fitted coordinate change, while cross-network representation-fit quality
is evaluated separately. We further measure the alignment's
underdetermined complement and show that it does not account for the
downstream identification result. The dominant low-dimensional spectral
core is already largely present before collapse completes, while
collapse primarily suppresses a low-energy tail. Within the aligned
support, donor-specific functional fingerprints remain distinguishable
after recipient-level baseline correction: all 20 ordered
donor-recipient pairs are correctly identified, with an exact
permutation p=0.0083, the resolution floor of the five-donor design.
Identification margins remain essentially unchanged after an
implementation leakage was identified, corrected, and the full analysis
re-run. These findings establish detectability under the operational
test used here, but not transplantability or causal persistence, which
require a separate intervention. More broadly, the study illustrates
how coordinate alignment, ambiguity diagnostics, leakage control, and
held-out identification can be combined to test for residual
cross-network functional variation in a controlled setting. Whether the
same protocol remains informative in other architectures, tasks, or
cross-model applications is an open empirical question.\end{abstract}

\section{Introduction}

\subsection{Motivation: the gauge problem in cross-network comparison}

Comparing independently trained neural networks is central to several
active lines of research. Studies of convergent learning ask whether
separate training runs discover common representations; model stitching
and model merging ask whether components from different networks can be
made compatible; and mechanistic interpretability increasingly compares
features or circuits across independently trained models. These
questions differ in their objectives and interventions, but they share
a basic measurement difficulty: independently trained networks do not
possess a common neuron-index reference frame.

A unit, direction, or classifier weight in one network therefore has no
immediate coordinate-wise correspondence to the same-indexed object in
another. Independently trained solutions may represent similar
computations using differently ordered or rotated internal coordinates,
while genuinely different functions may appear superficially similar
under aggregate geometric summaries. Without explicitly accounting for
this coordinate freedom, a cross-network discrepancy cannot be cleanly
interpreted as either meaningful functional variation or arbitrary
basis mismatch.

\subsection{The central question}

Neural Collapse provides a particularly controlled setting in which to
study this problem. Independently trained networks converge toward a
shared, low-dimensional geometric organization \citep{papyan2020prevalence}: within-class
variability contracts, class means approach an equiangular simplex, and
classifier directions align with that simplex. This common structure
removes one possible source of gross representational variation, but it
does not by itself establish that the resulting networks are
functionally indistinguishable.

Indeed, Neural Collapse sharpens the gauge problem rather than resolving
it. Once a shared geometry has emerged, any remaining cross-trajectory
difference may reflect at least three possibilities: unresolved
coordinate freedom, recipient-specific structure, or a donor-specific
functional residual. Distinguishing these possibilities requires more
than observing representational similarity. It requires an alignment
whose functional meaning is verified, a direct audit of its
underdetermined directions, and an identification test constructed
without using the evaluated query to define its own baseline.

This leads directly to the question addressed in this paper: when does
cross-trajectory variation reflect a detectable donor-specific
functional signature, and when can it be explained by coordinate gauge
or recipient-level structure? We take care to distinguish three
claims that this question can otherwise collapse into the single word
"identity." A donor-specific signature may be detectable, meaning
recoverable by a specific measurement procedure given access to both
networks. Transplantable, meaning capable of being moved into a
different network and remaining present there. Causally persistent,
meaning capable of continuing to influence a recipient network's
behavior after such a transplant, rather than being immediately
overwritten or absorbed. These three claims do not imply one another,
and a result at one level should not be read as evidence for the
others. This paper addresses the first. The remaining two define the
boundary of the paper rather than its conclusions.

\subsection{Approach overview}

We address this question through a deliberately constrained empirical
study. Five independently initialized MLP-5 networks were trained on
MNIST under a paper-matched reconstruction of a published two-phase
protocol, reported to induce Neural Collapse; this reconstruction is not
an exact replication, since some training details are not recoverable
from the published description, and we flag this explicitly rather than
assume it away. Because independently trained networks share no
coordinate frame, any comparison between them first requires an
alignment procedure; we use an orthogonal Procrustes fit combined with
an affine correction, verified to preserve the donor logits under the
fitted coordinate transformation exactly, while the quality of the
cross-network representation fit was evaluated separately. We then
characterize the shared low-dimensional structure of the aligned
representations before asking whether donor-specific functional
signatures remain distinguishable within that structure after
recipient-level baseline correction.

\subsection{Summary of findings}

Our findings form a sequence of five claims, each building on the
preceding one. Neural Collapse reproduces across all five seeds in the
tested regime, with collapse thresholds and feature norms falling within
a consistent range. The affine head transformation is verified to
preserve donor logits exactly under the fitted coordinate change, while
the quality of the cross-network representation fit is evaluated
separately. The specific underdetermined component examined here -- the
alignment's action outside the calibration-supported subspace -- is
measured rather than assumed and is shown not to account for the later
identification result. The residual structure remaining after alignment
lies within a shared, low-dimensional support that is already present
before collapse completes; we report this as a secondary structural
observation rather than the paper's central contribution. Within that
shared support, donor-specific functional fingerprints remain reliably
distinguishable after recipient-level baseline correction, at the
statistical resolution limit of a five-seed design, with identification
margins remaining essentially unchanged after an implementation leakage
was identified, corrected, and the full analysis re-run. Together, these
results establish that donor-specific functional fingerprints are
detectable under the operational test used here. They do not establish
that such fingerprints are transplantable or causally persistent under
intervention; no experiment in this paper moves structure between
networks or follows its fate after training resumes.

\subsection{Contributions}

This paper makes four contributions.

First, a leakage-audited protocol for testing donor-specific functional
detectability in Neural-Collapsed representations. The protocol
combines an affine-correct alignment procedure, explicit measurement of
the alignment's underdetermined complement, rank-based structural
characterization, disjoint calibration/template/probe splits, and
template-matching identification with an exact permutation test.
Together, these components provide a reproducible operational test for
distinguishing donor-specific signal from the coordinate and
data-dependence confounds examined in this setting. Whether the same
protocol transfers effectively to other architectures, tasks, or
cross-model applications remains to be tested.

Second, empirical evidence that donor-specific functional fingerprints
remain detectable after coordinate-gauge removal and recipient-level
baseline correction in the tested regime. All 20 ordered
donor-recipient pairs are correctly identified under disjoint template
and probe splits, with an exact permutation p-value at the resolution
limit of the five-donor design. This establishes donor-specific
detectability in the operational sense measured here; it does not
establish transplantability or causal persistence.

Third, a secondary structural observation: on the seeds tested, the
dominant low-dimensional spectral core Neural Collapse converges toward
is largely present before collapse completes, with the collapse phase
acting primarily to suppress a low-energy spectral tail rather than to
construct that core from scratch. This finding is reported at the scope
it was measured and is not advanced as a general account of Neural
Collapse.

Fourth, an explicit conceptual distinction -- detectable, transplantable,
and causally persistent identity -- intended to prevent the first of
these from being read as evidence for the other two, in this paper and
in future work addressing the same question.

The ordering above reflects the paper's own emphasis: the first two
contributions are central, the third is secondary, and the fourth is a
conceptual tool that applies to all of them.

The measurement problem motivating this study is not unique to Neural
Collapse. Questions about representation similarity, convergent
learning, model stitching, model merging, and cross-model
interpretability all require some account of how arbitrary coordinate
differences are separated from meaningful variation. The present paper
does not test those applications. Instead, it develops and audits one
operational procedure in a deliberately constrained Neural Collapse
setting. This setting provides a controlled case in which coordinate
gauge, shared low-dimensional geometry, and donor-specific residual
variation can be examined separately. Whether the resulting procedure
is useful beyond this setting is a direction for subsequent validation
rather than a conclusion of the present work.

\subsection{Roadmap}

Methods describes the training protocol, the alignment and rank-analysis
procedures, and the fingerprint-identification pipeline in full,
including the leakage found and corrected during development. Results
reports the reproduction of Neural Collapse across seeds, the
verification of the alignment procedure, the structural characterization
of the aligned representations, and the donor-identification results.
Discussion interprets these findings against the detectability/
portability distinction introduced above, situates the secondary
structural finding, states the paper's limitations directly, and
specifies the causal intervention required to move beyond detectability.
The paper is organized to separate implementation, evidence, and
interpretation as explicitly as possible.

\section{Methods}

\subsection{Experimental setup}

\subsubsection{Architecture and training protocol}

We trained an MLP-5 network (Flatten -> five Linear-ReLU blocks of width
512 -> linear classifier head, K=10) on MNIST, following a paper-matched
reconstruction of the two-phase protocol described by \citet{rupa2026neural}: a cross-entropy pretraining phase (200 epochs)
inducing high train accuracy, followed by an MSE phase (400 epochs,
one-hot targets) inducing Neural Collapse. Linear layers use
Kaiming-normal weight initialization with zero bias, matching the
architecture description in the source paper. Optimization used Adam
(lr=1e-3, weight decay=1e-4) with cosine annealing over the full
600-epoch run. Batch size (128) and the exact cosine-schedule horizon at
the Phase 1/2 transition are not specified in the source paper's
description and could not be verified against the original
implementation; both are stated here as provisional configuration choices
rather than confirmed replication details. We therefore describe our
reconstruction as paper-matched rather than an exact replication (see
Results 1).

\subsubsection{Seeds and checkpointing}

Five independent seeds (0-4) were trained under this identical protocol,
differing only in the random seed controlling weight initialization and
data-loader shuffling order. For each seed, we saved two checkpoints: the
end of the cross-entropy phase (Phase-1-end, epoch 200), and the first
epoch during the MSE phase at which NC1 crosses below 0.01 (T\_NC), the
threshold used throughout this paper to mark the onset of Neural
Collapse. Training and evaluation metrics (NC1, NC2, NC3, mean feature
norm, train accuracy) were logged every epoch. Deterministic seeding was
enforced for Python, NumPy, PyTorch (CPU and CUDA), and all data-loader
workers. To verify implementation reproducibility, seed 0 was
independently re-run in full and confirmed to reproduce its own
trajectory to the reported decimal precision at multiple checkpoints
(epoch 0, 199, 360, 599), including its T\_NC value and fn* to six
significant figures.

\subsubsection{Neural Collapse metrics}

NC1, NC2, and NC3 were computed following the definitions adopted by
\citet{rupa2026neural} and the Neural Collapse literature \citep{papyan2020prevalence}: NC1 measures within-class
variability collapse, NC2 measures simplex equiangular tightness, and NC3
measures self-duality between the classifier and the class means. We
used the sample-weighted global mean convention for the between-class
covariance. An alternative, class-balanced convention was also
implemented and compared; the two conventions differed by 0.5-0.6\% on
real checkpoint data across the metrics used in this paper, indicating
that the choice of global-mean convention was not a material contributor
to the findings reported here. Mean feature norm (fn) is computed as the
mean L2 norm of penultimate-layer activations across the evaluation set.

\subsection{Activation alignment}

\subsubsection{Affine-correct construction}

To compare a donor network B's classifier head with a recipient network
A's activations, we align B's penultimate-layer representation into A's
coordinate system via an orthogonal Procrustes fit \citep{schonemann1966generalized}. Given centered
calibration activations X\_A, X\_B in R\textasciicircum{}(n\_cal x d) (rows correspond to
examples, columns to activation dimensions), mean-subtracted using each
network's own calibration mean (mu\_A, mu\_B), we solve

\[Q^* = \underset{Q^\top Q = I}{\arg\min}\ \|X_B Q - X_A\|_F\]

via singular value decomposition. The donor head (weight W\_B, bias b\_B)
is then mapped into recipient coordinates as

\[\widetilde{W}_B = W_B Q, \qquad \widetilde{b}_B = b_B + W_B \mu_B - \widetilde{W}_B \mu_A,\]

which reproduces the donor's own logits on the donor's own activations
exactly under this construction (verified in Results 2). All subsequent
references to the "mapped donor head" refer to the pair (W\_tilde\_B,
b\_tilde\_B), never to the transformed weights alone. We refer to this
combined operation -- Procrustes fit followed by the affine correction
above -- as the affine-correct alignment throughout this paper.

\subsubsection{Calibration and held-out data}

Calibration activations were drawn from a class-balanced subset of the
MNIST training set, held disjoint from a fixed, class-balanced held-out
set (2000 examples) reserved for all downstream evaluation. Four
calibration sizes were tested (n\_cal in {100, 500, 1000, 5000}); the two
smallest are labeled stress tests, since d=512 exceeds n\_cal for these
sizes, and primary inference throughout this paper uses n\_cal=1000 or
5000 unless otherwise noted. The same calibration, template, and probe
partitions were reused across all alignment methods and seed pairs.

\subsubsection{Null-space diagnostics and completion sampling}

Because n\_cal can be smaller than the ambient activation dimension
(d=512), the Procrustes fit determines Q only on the subspace spanned by
the calibration activations; its action on the orthogonal complement is
formally unconstrained by the fit. For each fit, we computed the
numerical rank of the calibration activations (fraction of singular
values exceeding a fixed threshold, tau=1e-4 x the largest singular
value) to quantify how much of the d=512-dimensional space the
calibration set actually determines.

To test whether this unconstrained direction affects downstream logits,
we constructed a completion sampler that holds Q's action on the
determined subspace fixed while randomizing its action on the complement,
subject to the sampler itself remaining orthogonal. The original
implementation preserved the action of Q on vectors inside the supported
subspace, but did not preserve the projected map P\_B Q, which is the
quantity that determines the classifier's action after alignment and
therefore the quantity that must remain fixed for completions to be
functionally equivalent. This was detected via a direct numerical check,
corrected, and re-verified: orthogonality and P\_B-invariance were both
confirmed to hold to \textasciitilde{}1e-6 under the corrected construction (Table~\ref{tab:table2}). Completion sensitivity was then measured as the relative
standard deviation of held-out logits across multiple random completions
of a fitted Q.

\subsubsection{Rank battery}

For any activation matrix, we report four complementary rank statistics,
computed from its singular value spectrum sigma\_1 >= sigma\_2 >= ...: the
thresholded numerical rank at tau=1e-4 (and, as a robustness check, at
tau=1e-5 and 1e-6); the cumulative-energy ranks r\_90, r\_95, r\_99, and
r\_99.9 (the smallest number of components whose squared singular values
capture that fraction of total energy); and the effective rank, computed
as the participation ratio,

\[r_{\mathrm{eff}} = \frac{\left(\sum_i \sigma_i^2\right)^2}{\sum_i \sigma_i^4},\]

following \citet{gao2017theory}.

The effective rank is threshold-free and distinguishes a sharp low-rank
manifold from an activation spectrum that merely decays quickly without a
hard cutoff.

\subsection{Phase-1 vs. T\_NC rank comparison}

For each of the three seeds (0-2), we applied the rank battery defined in
Methods 2 to activations extracted at two checkpoints from the same
training trajectory: Phase-1-end and T\_NC. This is a within-seed
longitudinal comparison, not a cross-seed comparison: both checkpoints
being compared belong to the same network at two points in its own
training, so no alignment or cross-network mapping is involved. Unless
otherwise stated, the primary comparison reported in Results uses the
n\_cal=1000 calibration subset defined in Methods 2.

To determine whether the \textasciitilde{}13\% residual reconstruction error identified
in Results 2 lives inside or outside the shared supported subspace, we
re-expressed the donor-side supported-subspace projector P\_B in the
recipient's coordinate frame as P\_{B->A} = Q\textasciicircum{}T P\_B Q, and decomposed
held-out recipient activations into components parallel and
perpendicular to this mapped projector:

\begin{align*}
X_A &= X_A^{\parallel} + X_A^{\perp}, \\
X_A^{\parallel} &= X_A P_{B \to A}, \qquad X_A^{\perp} = X_A (I - P_{B \to A}).
\end{align*}

Using the donor-coordinate projector P\_B directly against
recipient-coordinate quantities would mix two different coordinate
frames; the mapped projector is required for this decomposition to be
valid, and its consistency with the donor-coordinate quantities it is
derived from was verified numerically as part of the alignment
verification described in Table~\ref{tab:table2}.

\subsection{Donor fingerprint identification}

\subsubsection{Additive donor-recipient decomposition}

To determine whether the pairwise alignment error found in Results 2 and 3
reflects donor-specific structure, recipient-specific structure, or both,
we fit an additive model to the 5x5 matrix of pairwise reconstruction
errors E\_ij (donor i, recipient j, i != j):

\[E_{ij} = \mu + \alpha_i + \beta_j + \gamma_{ij},\]

with sum-to-zero constraints on the donor effects {alpha\_i} and recipient
effects {beta\_j} for identifiability, fit by least squares to the 20
off-diagonal entries (the diagonal i=j is not a valid alignment
observation and is excluded). The residual term gamma\_ij is not modeled
further and is carried forward as the unexplained pair-specific
component. We report the model's R\textasciicircum{}2 and a leave-one-pair-out prediction
error (refitting with each pair held out in turn) as a check against
overfitting a 20-point matrix with 9 free parameters.

\subsubsection{Fingerprint construction}

For each ordered (donor, recipient) pair, we constructed a fingerprint
summarizing the mapped donor head's (Methods 2) behavior on recipient
activations. Given logits L in R\textasciicircum{}(n x K) on some example set with true
labels y, the fingerprint is

\begin{align*}
M_c &= \text{mean over } \{i : y_i = c\} \text{ of } (L_i - \text{mean}_k\, L_{i,k}), \quad c = 1 \ldots K, \\
Z &= M / \|M\|_F,
\end{align*}

a K x K matrix (Z in R\textasciicircum{}(K x K)) whose row c is the class-centered mean
logit vector for class c, normalized to unit Frobenius norm. Centering is
performed independently for each example (subtracting that example's own
across-class mean logit) before class averaging, not the reverse.
Anchoring the fingerprint to class identity, rather than to a fixed set
of specific example indices, allows two fingerprints built from entirely
disjoint sets of images to be compared directly: class c means the same
thing in both, even when the specific images differ, whereas a raw
per-example fingerprint would not be comparable across disjoint example
sets. The distance between two fingerprints is
D(Z, Z') = 1 - cos(vec(Z), vec(Z')).

Each donor's template fingerprint was built from a fixed template split
(1000 examples, class-balanced, disjoint from calibration); each query
fingerprint was built from a disjoint probe split (1000 examples,
class-balanced, disjoint from both calibration and template). Both splits
are drawn from the same fixed 2000-example held-out pool defined in
Methods 2.

\subsubsection{Recipient-level baseline correction}

Because Section 4.1 shows the raw pairwise error matrix contains a
substantial additive recipient-level component, we additionally test
donor identification after removing a recipient-level baseline from each
fingerprint. For a given recipient A with donor candidates {B\_1,...,B\_m},
the baseline is the across-donor mean of the TEMPLATE fingerprints at A:

\[\mathrm{baseline}_A = \text{mean over } k \text{ of } Z(\text{template for donor } B_k \text{ at recipient } A),\]

fit using the template split only. This same baseline is then subtracted
from both the template and the probe fingerprints before matching. An
earlier implementation instead computed a separate baseline for the probe
split, using the across-donor mean of the probe fingerprints themselves,
which allowed the probe fingerprint currently being evaluated to
contribute to its own baseline (a data-dependence we verified directly:
the baseline used in that version changed when the probe fingerprint
being evaluated changed, even though no other probe fingerprint changed).
This was identified, corrected to the template-only baseline correction, and the identification pipeline was re-run in full; we verified
numerically that the corrected baseline is invariant to any change in
probe content. All reported Stage 0D results were produced using the
leakage-corrected implementation described in this section; Results 4 reports both
the original and corrected numbers for direct comparison.

\subsubsection{Identification rule and margin}

A query fingerprint is assigned to the donor whose template fingerprint
is nearest under D(.,.). For each true donor B at recipient A, we define
the identification margin as

\[\mathrm{margin} = \min_{B' \neq B} D(\mathrm{query}_B, \mathrm{template}_{B'}) - D(\mathrm{query}_B, \mathrm{template}_B),\]

the distance to the nearest incorrect donor's template minus the distance
to the correct donor's template. A positive margin indicates a correct
identification; negative margins correspond to misidentification. The
magnitude of the margin relative to the overall scale of pairwise
distances indicates how robust that identification is to small
perturbations, as distinct from the binary correct/incorrect outcome
alone.

\subsubsection{Exact donor-level permutation test}

Statistical significance of the identification accuracy was assessed by
an exact permutation test over donor identity labels, applied jointly
across all five recipient folds rather than independently within each
fold. Because each donor's fingerprint is evaluated inside multiple
recipient contexts, per-fold identity labels are not independent of one
another; the test therefore enumerates all 5! = 120 permutations of the
five donor labels, applies each permutation sigma consistently across
every recipient fold's ground truth, and recomputes overall accuracy
under that joint relabeling. A relabeled donor identity that would equal
the recipient itself (an invalid donor-recipient pair, since a network is
never its own donor) remains part of the permutation enumeration but
necessarily contributes zero correct matches. The reported p-value is
the fraction of the 120 permutations attaining accuracy at least as high
as the accuracy actually observed; because only the identity permutation
can achieve the maximum observed accuracy in our results, p=1/120=0.0083
is the minimum value this exact test can attain with five donor labels,
not a chosen or Monte-Carlo-estimated figure.

\section{Results}

This section proceeds in four stages. We first establish a reproducible
Neural Collapse baseline, then validate the alignment procedure used for
cross-network comparison, characterize the resulting shared
representation, and finally test whether donor-specific fingerprints
remain detectable after controlling for recipient-level structure.

\subsection{Neural Collapse dynamics reproduce across independent seeds}

We first establish that the checkpoints used throughout the remainder of
the paper correspond to genuine Neural Collapse states rather than
artifacts of a particular training run. Five independently initialized
MLP-5 networks were trained on MNIST under the two-phase protocol of
\citet{rupa2026neural} (cross-entropy pretraining followed by an MSE phase inducing Neural
Collapse), and NC1, NC2, NC3, and mean feature norm (fn) were tracked
throughout Phase 2 (Fig. 1). All five seeds cross the NC1 collapse
threshold (NC1 < 0.01) within a narrow window: T\_NC ranges from 335 to
360 epochs (Table~\ref{tab:table1}), and the feature norm at collapse, fn*, ranges from
0.889 to 0.952 — both consistent with the regime reported by \citet{rupa2026neural}. This
is a paper-matched reconstruction rather than an exact replication. Exact
replication cannot be claimed because batch size and the precise
scheduler behavior at the Phase 1/2 transition are not recoverable from
the published description alone; we state this as an explicit limitation
rather than assume it away.

\begin{table}[htbp]
\centering
\caption{Neural Collapse baseline across five independent seeds. $T_{NC}$ reported as global-epoch / Phase-2-relative-epoch; fn* is the mean feature norm at the $T_{NC}$ checkpoint; final acc is train accuracy at epoch 599.}
\label{tab:table1}
\begin{tabular}{ccccc}
\toprule
seed & $T_{NC}$ (global / Phase-2) & fn* & final train acc \\
\midrule
0 & 360 / 160 & 0.930 & 0.9993 \\
1 & 335 / 135 & 0.952 & 0.9994 \\
2 & 353 / 153 & 0.922 & 0.9993 \\
3 & 349 / 149 & 0.910 & 0.9994 \\
4 & 340 / 140 & 0.889 & 0.9994 \\
\bottomrule
\end{tabular}
\end{table}

\begin{table}[htbp]
\centering
\caption{Post-collapse rebound in NC2/NC3, per seed. Sustained minimum computed via a 5-epoch median window; gap and rebound percentages measured relative to that minimum.}
\label{tab:table1b}
\resizebox{\textwidth}{!}{%
\begin{tabular}{cccc}
\toprule
seed & gap: min $\to T_{NC}$ (NC2/NC3, epochs) & rebound at $T_{NC}$ (NC2/NC3) & rebound at final epoch (NC2/NC3) \\
\midrule
0 & 44 / 34 & +2.1\% / +2.0\% & +11.3\% / +18.7\% \\
1 & 106 / 85 & +12.2\% / +10.5\% & +23.9\% / +22.0\% \\
2 & 90 / 90 & +1.8\% / +1.1\% & +5.9\% / +5.7\% \\
3 & 40 / 48 & +1.1\% / +1.8\% & +10.9\% / +20.4\% \\
4 & $\sim$0 / 1 & +0.0\% / -0.4\% & +6.4\% / +4.7\% \\
\bottomrule
\end{tabular}%
}
\end{table}

\begin{figure}[htbp]
\centering
\includegraphics[width=0.85\textwidth]{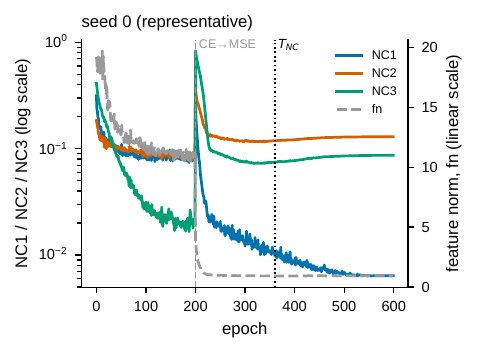}
\caption{Neural Collapse dynamics across training, seed 0 (representative). NC1, NC2, and NC3 (left axis, log scale) and mean feature norm fn (right axis, linear scale) are tracked per epoch. The dashed gray line marks the Phase 1 to Phase 2 transition (epoch 200); the dotted black line marks $T_{NC}$ (epoch 360 for seed 0). NC2 and NC3 reach a sustained minimum before $T_{NC}$ and partially rebound afterward (Table~\ref{tab:table1b}), visible directly in the trajectory shown here.}
\label{fig:fig1}
\end{figure}

Beyond the NC1 threshold itself, we observe a secondary,
unanticipated phenomenon: NC2 and NC3 do not decrease monotonically to
their eventual floor. Instead, each trajectory reaches a sustained minimum
and then partially rebounds over the following hundreds of epochs
(Table 1b). This rebound is directional and consistent across all five
seeds — every seed's final NC2 and NC3 values exceed their own sustained
minimum. The magnitude and timing of this rebound, however, differ
substantially across seeds. In four of five seeds (0, 2, 3, 4), T\_NC
falls close to the sustained minimum (within 0–90 epochs, and within 2\%
of the minimum value); in one seed (1), the rebound is already
substantial (10–12\%) by the time NC1 crosses threshold. We report this
pattern descriptively and do not offer a mechanistic account for seed 1's
departure from the other four. The downstream analyses therefore use the
NC1-threshold checkpoint itself, rather than assuming that NC2 and NC3
have simultaneously reached their own minima.

Together, these results represent reproducible collapsed checkpoints
suitable for all subsequent analyses.

\subsection{Activation alignment removes coordinate gauge while preserving donor function}

Independent neural networks do not share a neuron-index reference frame.
Consequently, any direct comparison of independently trained
representations is ill-defined unless both networks are first expressed
in a common coordinate system. We therefore validated the
activation-alignment construction used throughout this paper before
using it to ask any question about donor-specific structure.

The alignment maps a donor network's classifier head into a recipient
network's coordinate system via an orthogonal Procrustes fit on shared
calibration activations (Fig. 2a). We first verified that this
construction is exact: applied to a donor's own activations, it
reproduces the donor's own logits to floating-point precision (relative
error 1.3e-15). The Procrustes fit is well-determined only on the
subspace spanned by the calibration activations, leaving its action on
the orthogonal complement formally unconstrained. We built a completion
sampler that varies this unconstrained action while holding the
well-determined part fixed, verified the sampler's correctness on
synthetic data (Fig. 2b), and confirmed a synthetic extreme case: when a
donor head is confined exactly to the well-determined subspace,
completion sensitivity is at machine precision, exactly as required.

Despite exact recovery of the donor function under the affine
construction, held-out logit reconstruction remained imperfect
(approximately 13\% relative error). The remaining question is whether
this error is an artifact of the underdetermined coordinate directions,
or whether it reflects genuine structure within the part of the
representation the alignment has already determined. Completion
sensitivity measured directly on the T\_NC checkpoints is effectively
zero, ruling out the first possibility.

The remaining reconstruction error must therefore arise within the shared
supported subspace itself. We next characterize that shared subspace,
then ask whether the remaining variation within it carries donor-specific
information.

\begin{table}[htbp]
\centering
\caption{Alignment verification. Rows 1-3: synthetic/numerical checks of the completion-sampler construction. Row 4: the same completion-sensitivity test applied directly to real $T_{NC}$ checkpoints (seeds 0-2, all 6 ordered pairs).}
\label{tab:table2}
\begin{tabular}{p{5.2cm}p{3cm}p{5.5cm}}
\toprule
Verification & Result & Interpretation \\
\midrule
Orthogonality ($\|Q_c^\top Q_c - I\|$) & $\sim 10^{-6}$ & valid orthogonal completion \\
Supported-subspace invariance ($\|P_B Q_c - P_B Q_{\mathrm{fit}}\|$) & $\sim 10^{-6}$ & donor-side action preserved across completions \\
Extreme-case completion sensitivity (synthetic; donor head confined exactly to $P_B$) & $2.9 \times 10^{-6}$ & null-space ambiguity is functionally harmless when it should be, by construction \\
Real-data completion sensitivity ($T_{NC}$ checkpoints, seeds 0-2, 6 ordered pairs) & relative sensitivity $\sim 0.0000$; sampler audit passed; max $\Delta_{\mathrm{supported}} \sim 10^{-5}$ & null-space ambiguity does not drive the $\sim$13\% residual reconstruction error observed in Results 2 \\
\bottomrule
\end{tabular}
\end{table}

\begin{figure}[htbp]
\centering
\includegraphics[width=\textwidth]{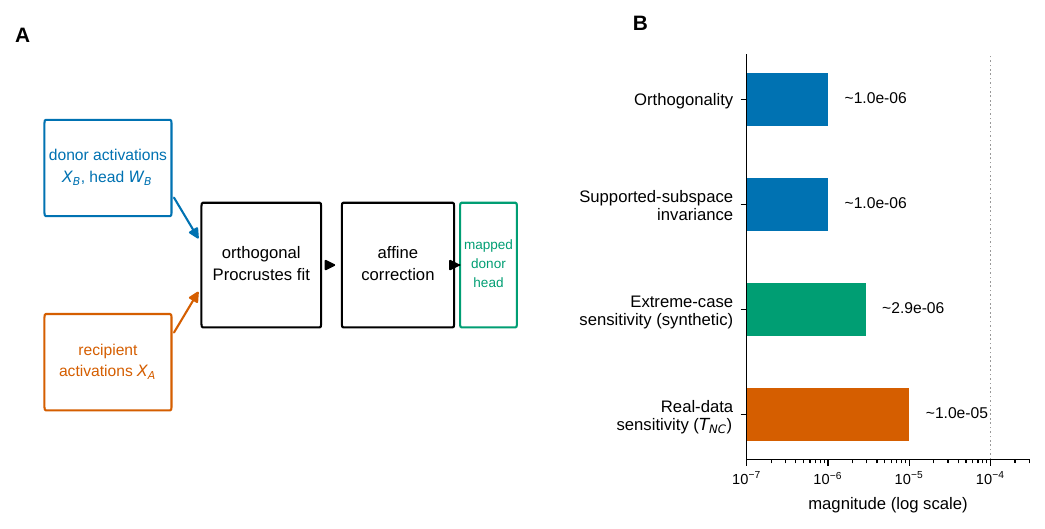}
\caption{The affine-correct alignment construction and its verification. (A) Donor activations and classifier head are mapped into the recipient's coordinate frame via an orthogonal Procrustes fit followed by an affine correction. (B) Four verification checks confirm the construction is valid (Table~\ref{tab:table2}), all several orders of magnitude below the $\sim$13\% held-out reconstruction error reported in this section.}
\label{fig:fig2}
\end{figure}

\subsection{Neural Collapse selectively removes a low-energy spectral tail while preserving a pre-existing low-dimensional core}

The alignment analysis above establishes a shared coordinate system but
does not explain why the aligned representations occupy such a
low-dimensional support. To address this, we compared the same three
checkpoints at two points in training — the end of Phase 1, before
Neural Collapse begins, and T\_NC, the moment NC1 crosses threshold.

\begin{figure}[htbp]
\centering
\includegraphics[width=\textwidth]{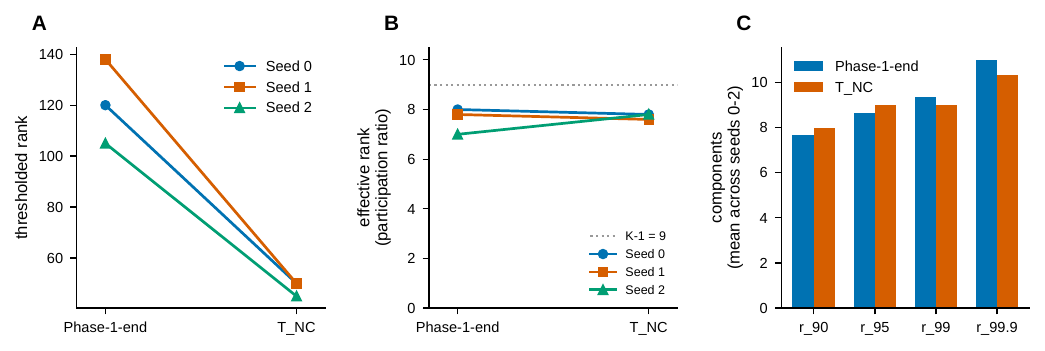}
\caption{Neural Collapse suppresses a low-energy spectral tail while
preserving a pre-existing low-dimensional core. (A) Thresholded numerical
rank decreases sharply from Phase-1-end to T\_NC in all three seeds. (B)
Effective rank, computed as the participation ratio, remains near K-1=9
across the same interval. Lines connect checkpoints from the same
training trajectory. The divergence between thresholded and effective
rank indicates that Phase 2 primarily removes weak spectral directions
rather than constructing the dominant low-dimensional core de novo. (C)
Cumulative-energy ranks (r\_90, r\_95, r\_99, r\_99.9), averaged across the
three seeds, show most activation energy already concentrated within
approximately 7-12 components at both checkpoints.}
\label{fig:fig3}
\end{figure}

The thresholded numerical rank drops sharply between these two
checkpoints, from 105–138 at Phase-1-end to 45–50 at T\_NC, across all
three seeds tested (Table~\ref{tab:table3}, Fig.~\ref{fig:fig3}). A single thresholded rank number, however,
can conflate a genuine low-dimensional manifold with an activation
spectrum that merely decays quickly without a sharp cutoff. We therefore
also computed effective rank (participation ratio), a threshold-free
measure of how many dimensions carry meaningful energy. Effective rank is
nearly unchanged between the two checkpoints — approximately 7.0 to 8.0
at both Phase-1-end and T\_NC — and close to K-1 = 9, the theoretical
ceiling implied by the equiangular structure Neural Collapse converges
toward.

The dominant activation spectrum is already concentrated into roughly
K-1 modes by the end of Phase 1. Phase 2 therefore acts primarily to
suppress a broad, low-energy spectral tail rather than to create the
low-dimensional core from scratch. Thus, Neural Collapse appears to act
primarily as a spectral-tail pruning process rather than constructing the
dominant low-dimensional representation de novo.

\begin{table}[htbp]
\centering
\caption{Phase-1-end vs. $T_{NC}$ rank battery (within-seed, seeds 0-2). Hard rank thresholded at $\tau = 10^{-4}$; effective rank is the participation ratio (Methods, ``Rank battery''). $K-1=9$ (10-class task), the theoretical ceiling implied by the equiangular structure Neural Collapse converges toward.}
\label{tab:table3}
\begin{tabular}{ccccc}
\toprule
seed & P1-end hard rank & $T_{NC}$ hard rank & P1-end effective rank & $T_{NC}$ effective rank \\
\midrule
0 & 120 & 50 & 8.0 & 7.8 \\
1 & 138 & 50 & 7.8 & 7.6 \\
2 & 105 & 45 & 7.0 & 7.8 \\
\bottomrule
\end{tabular}
\end{table}

\begin{table}[htbp]
\centering
\caption{Mapped-projector decomposition and null-completion diagnostics. $T_{NC}$ checkpoints, seeds 0-2, all 6 ordered (donor, recipient) pairs, using the corrected mapped projector $P_{B \to A} = Q^\top P_B Q$. $r$ = shared supported-subspace dimension. All 6 pairs: invariance audit passed; null-space completion sensitivity = 0.0000 in every case.}
\label{tab:table4}
\resizebox{\textwidth}{!}{%
\begin{tabular}{ccccccccc}
\toprule
donor & recipient & $r$ & invariance err & $\eta_{A_\perp}^{\mathrm{mapped}}$ & $\eta_{W_\perp}$ & $E_{\mathrm{logit,total}}$ & $|L_{\parallel}|$ & $|L_{\perp}|$ \\
\midrule
0 & 1 & 50 & $5.82\times10^{-11}$ & 0.000 & 0.000 & 0.1284 & 42.65 & 0.00 \\
0 & 2 & 50 & $1.16\times10^{-10}$ & 0.000 & 0.000 & 0.1397 & 42.35 & 0.00 \\
1 & 0 & 50 & $4.66\times10^{-10}$ & 0.000 & 0.002 & 0.1285 & 42.46 & 0.00 \\
1 & 2 & 50 & $1.16\times10^{-10}$ & 0.000 & 0.002 & 0.1243 & 42.21 & 0.00 \\
2 & 0 & 45 & $1.46\times10^{-10}$ & 0.000 & 0.000 & 0.1405 & 42.93 & 0.00 \\
2 & 1 & 45 & $2.91\times10^{-11}$ & 0.000 & 0.000 & 0.1293 & 43.02 & 0.00 \\
\bottomrule
\end{tabular}%
}
\end{table}

\begin{table}[htbp]
\centering
\caption{Calibration-resampling stability (5 seeds, 10 independent resamples, $n_{\mathrm{cal}}=1000$). Every pair: std $\le 0.0001$, CV $= 0.001$. Mean values shown; identical to four decimal places to the pairwise error matrix analyzed in Table~\ref{tab:table6}.}
\label{tab:table5}
\begin{tabular}{cccccc}
\toprule
donor $\backslash$ recipient & 0 & 1 & 2 & 3 & 4 \\
\midrule
\textbf{0} & -- & 0.1284 & 0.1398 & 0.1730 & 0.1614 \\
\textbf{1} & 0.1285 & -- & 0.1244 & 0.1712 & 0.1546 \\
\textbf{2} & 0.1407 & 0.1294 & -- & 0.1417 & 0.1373 \\
\textbf{3} & 0.1895 & 0.1949 & 0.1467 & -- & 0.1686 \\
\textbf{4} & 0.1671 & 0.1660 & 0.1390 & 0.1685 & -- \\
\bottomrule
\end{tabular}
\end{table}

We further confirmed that the \textasciitilde{}13\% residual reconstruction error
identified above lives within this shared, well-determined support
rather than outside it: after correcting the projector to the recipient's
coordinate frame, essentially none of a held-out network's activation
energy or reconstruction error falls in the unsupported complement
(Table 4). This \textasciitilde{}13\% residual is also stable: repeated recalibration on
independent draws of the calibration set changes it by less than 0.1\%
(CV = 0.001, Table 5).

Having ruled out coordinate gauge as the source of the remaining
reconstruction error, we now show that the error resides within a
shared, low-dimensional support. The remaining question is whether
variation inside this support carries donor-specific information.

\subsection{Donor-specific fingerprints remain distinguishable after recipient-level baseline correction}

\begin{figure}[htbp]
\centering
\includegraphics[width=\textwidth]{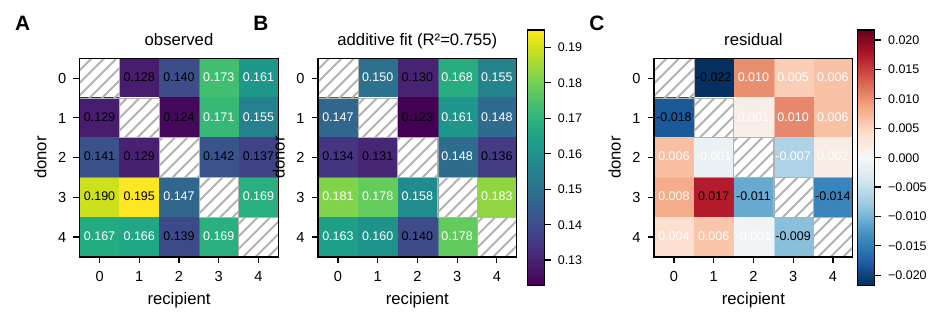}
\caption{Donor-specific fingerprints emerge after removing a large
additive donor-recipient component. (A) Observed pairwise
alignment-error matrix. (B) Additive reconstruction obtained from donor
and recipient main effects. (C) Residual matrix after subtracting the
additive component. The additive model explains 75.5\% of the pairwise
variance, leaving a structured residual that motivates recipient-level
baseline correction before donor identification (Results 4).}
\label{fig:fig4}
\end{figure}

The preceding analysis localizes a stable residual reconstruction error
to a shared, low-dimensional support, but does not determine whether
variation within that support is donor-specific. Because pairwise
alignment error reflects both donor- and recipient-specific effects, raw
distances alone cannot answer this question. An additive model fitted to
the 20 off-diagonal entries of the 5×5 donor–recipient matrix explained
75.5\% of their variance. Its leave-one-pair-out prediction error was 1.5
times the in-sample residual, consistent with a substantial additive
structure that generalizes beyond the fitted entries. This structure
motivated recipient-level baseline correction before donor fingerprint
identification (Table~\ref{tab:table6}, Fig.~\ref{fig:fig4}).

\begin{table}[htbp]
\centering
\caption{Additive donor-recipient decomposition. Fitted to the 20 off-diagonal entries of the pairwise error matrix in Table~\ref{tab:table5} via least squares with sum-to-zero constraints. $R^2 = 0.755$. Leave-one-pair-out mean absolute error $=0.0150$ (1.5$\times$ the in-sample residual mean absolute error).}
\label{tab:table6}
\begin{tabular}{ccc}
\toprule
seed & raw donor-row mean & raw recipient-column mean \\
\midrule
0 & 0.1507 & 0.1565 \\
1 & 0.1447 & 0.1547 \\
2 & 0.1373 & 0.1375 \\
3 & 0.1749 & 0.1636 \\
4 & 0.1602 & 0.1555 \\
\bottomrule
\end{tabular}
\end{table}

\begin{table}[htbp]
\centering
\caption{Donor-identification summary, raw vs. recipient-baseline-corrected. All 20 ordered donor-recipient pairs, 5 seeds. Exact permutation test: $5! = 120$ permutations, joint relabeling across all recipient folds. ``Pre-fix'' columns show the identification pipeline before the recipient baseline-leakage was corrected; ``Leak-free (primary)'' is the corrected, reported result. $p=0.0083=1/120$ is the minimum attainable value under this exact test with five donor labels (only the identity permutation matches the observed accuracy), not a chosen or estimated figure. The near-zero pre-fix correct-template distances (0.0000--0.0001) reflect the leakage identified and corrected during development; the leak-free distances (0.004--0.028) are the trustworthy values.}
\label{tab:table7}
\resizebox{\textwidth}{!}{%
\begin{tabular}{lccc}
\toprule
Metric & Raw (no correction) & Residualized, pre-fix (has leakage) & Residualized, leak-free (primary) \\
\midrule
Overall accuracy & 1.000 (20/20) & 1.000 (20/20) & 1.000 (20/20) \\
Exact permutation $p$ & 0.0083 & 0.0083 & 0.0083 \\
Correct-template distance, range & 0.0000--0.0001 & 0.0000--0.0001 & 0.004--0.028 \\
Minimum identification margin & 0.0055 & 0.8839 & 0.8618 \\
Median identification margin & 0.0061 & 1.1227 & 1.1187 \\
Mean identification margin & 0.0068 & 1.1206 & 1.1062 \\
Maximum identification margin & 0.0112 & 1.3084 & 1.2963 \\
\bottomrule
\end{tabular}%
}
\end{table}

\begin{table}[htbp]
\centering
\caption{Confusion matrices, raw and residualized (leak-free), pooled across all 5 recipient folds. Rows = true donor, columns = predicted donor. Both blocks are perfectly diagonal.}
\label{tab:table8}
\begin{minipage}{0.48\textwidth}
\centering
\textbf{Block A -- Raw identification}\\[4pt]
\begin{tabular}{cccccc}
\toprule
true $\backslash$ pred & 0 & 1 & 2 & 3 & 4 \\
\midrule
\textbf{0} & 4 & 0 & 0 & 0 & 0 \\
\textbf{1} & 0 & 4 & 0 & 0 & 0 \\
\textbf{2} & 0 & 0 & 4 & 0 & 0 \\
\textbf{3} & 0 & 0 & 0 & 4 & 0 \\
\textbf{4} & 0 & 0 & 0 & 0 & 4 \\
\bottomrule
\end{tabular}
\end{minipage}%
\hfill
\begin{minipage}{0.48\textwidth}
\centering
\textbf{Block B -- Residualized (leak-free)}\\[4pt]
\begin{tabular}{cccccc}
\toprule
true $\backslash$ pred & 0 & 1 & 2 & 3 & 4 \\
\midrule
\textbf{0} & 4 & 0 & 0 & 0 & 0 \\
\textbf{1} & 0 & 4 & 0 & 0 & 0 \\
\textbf{2} & 0 & 0 & 4 & 0 & 0 \\
\textbf{3} & 0 & 0 & 0 & 4 & 0 \\
\textbf{4} & 0 & 0 & 0 & 0 & 4 \\
\bottomrule
\end{tabular}
\end{minipage}
\end{table}

We constructed a fingerprint for each donor's function inside each
recipient's coordinate system, using per-class mean logit geometry on a
held-out probe set, and matched each query fingerprint to the nearest of
several candidate donor templates (built from a disjoint template split).
Raw fingerprint matching achieves 100\% identification accuracy across all
20 ordered donor-recipient pairs; an exact permutation test over all
possible donor-label relabelings (5! = 120) gives p = 0.0083, the minimum
attainable value with five donor labels because only the identity
relabeling matches the observed accuracy (Table~\ref{tab:table7}, Table~\ref{tab:table8}).

During implementation, we identified an earlier version in which the
recipient baseline was estimated using the query split being evaluated,
allowing each query to contribute to its own correction. This
implementation leakage was identified, isolated, corrected, and
re-evaluated before any conclusions were finalized. The range of
correct-template distances across all 20 queries increased from
approximately 0.0000–0.0001 to 0.004–0.028, confirming that the original
near-zero self-distances were artificial. Nevertheless, the minimum
identification margin changed only from 0.8839
to 0.8618, and the mean margin from 1.1206 to 1.1062. Accuracy remained
100\%, and the exact permutation p-value remained 0.0083 (Table~\ref{tab:table7}).

As an additional robustness check against a simpler explanation, we
repeated the donor-identification analysis using a fingerprint
constructed solely from per-class accuracy, while keeping the
alignment, calibration, template/probe splits, and evaluation protocol
unchanged. This baseline identified donors at chance level (25\%, exact
permutation $p=0.33$), with identification margins centered near zero.
Thus, per-class accuracy alone does not explain the donor-identification
result reported here. This robustness check does not rule out all
performance-related explanations; it addresses only the specific
hypothesis that donor identification is driven solely by per-class
accuracy.

Donor-specific fingerprints therefore remain distinguishable after
recipient-level baseline correction. This provides detectability
evidence that the aligned representation retains donor-specific
information sufficient for reliable identification. It does not
establish portability in the stronger sense of a fingerprint that can be
transplanted and shown to causally influence subsequent dynamics; that
question requires a separate intervention. The Discussion returns to this
distinction between detectability and portability, and considers what
additional causal evidence would be required to bridge it.
\section{Discussion}

\subsection{Identity beyond gauge}

The results presented above address the central question motivating this
work: when does cross-trajectory variation reflect a detectable
donor-specific functional signature, and when can it be explained by
coordinate gauge or recipient-level structure? We first removed
coordinate mismatch through a verified affine-correct alignment (Results
2), then further corrected for a large additive recipient-level
component in the resulting error structure (Results 4). After both
steps, donor-specific functional fingerprints remain reliably
identifiable: all 20 ordered donor-recipient pairs are correctly
matched, with an exact permutation test at the resolution floor of the
five-donor design (p=0.0083) and identification margins that are large
relative to the overall distance scale. An additional robustness
analysis showed that replacing the proposed fingerprint with a
per-class-accuracy fingerprint reduced identification to chance level
(Results 4), supporting the interpretation that the reported signal is
not attributable to class-wise accuracy alone.

This finding should be read at the level it was measured. What has been
shown is that a specific, operational procedure -- per-class mean logit
geometry, matched against held-out templates -- reliably distinguishes
one donor network's aligned function from another's. This is identity in
an operationally detectable sense: a signature recoverable by
measurement. It is not a claim about what a trained network "is" in any
deeper sense, and it is not yet a claim that this signature can be moved
into another network and continue to shape its behavior. The distinction
between detectability and portability therefore becomes the central
conceptual question for the remainder of this Discussion.

\subsection{Why this is not merely representational gauge}

A natural objection is that a sufficiently flexible cross-network
transformation can either manufacture or erase apparent differences, so
successful identification alone does not establish that the detected
signal lies beyond coordinate gauge. Three properties of our
construction address this directly, in sequence.

First, the comparison never relies on raw neuron-index correspondence
between networks. Independent networks have no shared neuron ordering,
so any procedure that compared activations coordinate-by-coordinate
without first aligning them would be comparing independently
parameterized bases with no justified neuron-wise correspondence
(Results 2).

Second, we used a restricted, orthogonal Procrustes transformation rather
than an unconstrained map, together with the affine correction required
to transform the classifier head consistently. The affine construction
was verified algebraically and numerically: under the fitted coordinate
transformation, the mapped head preserves the donor logits to
floating-point precision. This verification establishes that subsequent
discrepancies arise from the cross-network representation fit, rather
than from an inconsistent transformation of the classifier weights or
bias. It does not imply that the two networks' activations are matched
exactly; held-out reconstruction remains imperfect, as reported in Results 2.

Third, that alignment has a genuinely underdetermined component -- its
action on the subspace orthogonal to the calibration data. Under the
completion-sensitivity test used here, this ambiguity did not account for
the result: sensitivity on real checkpoints was effectively zero, and the
residual error the identification procedure operates on was confirmed to
live inside the well-determined part of the alignment, not the ambiguous
part (Results 2 and 3).

Taken together, these three properties locate the donor-specific signal
inside the part of the representation that alignment has actually
determined, not in an unconstrained rotational degree of freedom, and not
in an artifact of an exactly-matched representation fit that does not in
fact exist. This does not mean every conceivable form of coordinate gauge
has been ruled out; it means the specific gauge freedom present in this
construction was measured and shown not to be responsible for the
result.

\subsection{Detectability versus portability}

The analyses above show that donor-specific fingerprints remain reliably
identifiable after coordinate-gauge removal and recipient-level
correction. It is tempting to interpret this result as evidence for a
"portable trajectory identity": a signature that survives transfer from
one network to another and continues to shape the recipient's behavior.
That reading would overstate what has been shown here, and we want to be
precise about why.

We distinguish three empirically separate claims that can otherwise be
compressed into the single word "identity":

- \textbf{Detectable identity}: a donor-specific signature is recoverable by a
  specific measurement procedure, given access to both networks.
- \textbf{Transplantable identity}: that signature, or the structure producing
  it, can be moved into a different network and remains present there.
- \textbf{Causally persistent identity}: once transplanted, that structure
  continues to influence the recipient network's behavior or trajectory
  under continued training, rather than being immediately overwritten or
  absorbed.

This paper establishes the first claim. Results 2 through 4 show that a
donor's function, once correctly aligned into a recipient's coordinate
frame and corrected for a shared recipient-level component, remains
distinguishable from other donors' functions with large margins and at
the statistical resolution limit of a five-seed design. No experiment in
this paper directly tests the second or third claim: we did not
transplant any donor structure into a recipient network, and we did not
continue training any network after such a transplant to observe whether
the donor signature persists, is redirected, or is erased. This is an
absence of direct evidence on those claims, not evidence of absence --
the experiments needed to address them were simply not run here.

These three empirical claims can come apart. A fingerprint may be
detectable yet not transplantable if it depends on distributed structure
that cannot be moved through the selected intervention. It may be
transplantable yet not causally persistent if resumed training rapidly
absorbs or overwrites it. Establishing detectability therefore provides
an operational reference for testing portability: it supplies a
measurable signature whose fate can be followed before and after
intervention, but does not establish that the signature can be
transferred or remain causally influential. Testing those claims requires
a separate intervention that transfers an aligned donor component and
follows both the fingerprint and the recipient dynamics after training
resumes.

\subsection{Neural Collapse as convergence without complete identity erasure}

The preceding results support two observations that should be
interpreted together rather than reduced to a single narrative. First,
independently trained networks converge, under Neural Collapse, toward a
shared, low-dimensional geometric structure: the dominant activation
spectrum is concentrated into approximately K-1 modes, and this
concentration is present by the end of pretraining rather than being
newly constructed during collapse (Results 3). Second, this convergence
does not erase all donor-specific functional variation: after accounting
for the shared structure and a large additive recipient-level component,
donor-specific fingerprints remain reliably distinguishable (Results 4).

Read together, these observations indicate that convergence to a common
geometric equivalence class is not the same as convergence to a single
functionally indistinguishable internal state. Networks can occupy
essentially the same low-dimensional representational geometry, within
the resolution of the measurements used here, while still differing in
ways that a sufficiently careful, gauge-corrected comparison can recover.

We want to be explicit about what this coexistence does and does not
imply. It does not indicate that either the shared geometry or the
donor-specific variation is the more fundamental or primary property of
the trained network; the present experiments only establish that both
are simultaneously measurable within the same set of checkpoints. We
therefore avoid framing this finding as either "Neural Collapse erases
identity" or "Neural Collapse preserves identity" -- both would claim
more than a conjunction of two measured facts supports. The more precise
statement is that shared structure and donor-specific structure coexist
in these checkpoints, inferred from two complementary measurement
procedures (Methods 2 and 3 for the former, Methods 4 for the latter), and
neither measurement displaces the other.

\subsection{Secondary finding: spectral-tail pruning}

In addition to the paper's central question, the Phase-1-versus-T\_NC
comparison in Results 3 yields a secondary observation that is worth
recording in its own right. On the three seeds tested, thresholded
numerical rank falls sharply between the end of pretraining and the
onset of Neural Collapse, while effective rank -- a threshold-free
measure of how many dimensions carry meaningful activation energy --
stays approximately constant across the same interval, close to the
K-1 value implied by the equiangular structure Neural Collapse converges
toward.

One interpretation of these observations is that the dominant,
energy-concentrated structure of the representation is already present
by the end of pretraining, and that the subsequent collapse phase acts
mainly to suppress a broad, low-energy spectral tail rather than to
construct the low-dimensional structure from scratch. We offer this as a
reading of the present reconstruction -- three seeds, one architecture,
one dataset -- not as a general mechanism of Neural Collapse. Whether the
same pattern holds across other architectures, tasks, or
collapse-inducing training regimes is a separate empirical question this
paper does not address.

We flag this finding explicitly as secondary. It is not required to
support, nor does it materially alter, the paper's central claim
regarding donor-specific fingerprints (Results 4); it is reported because
it bears on how the shared low-dimensional structure discussed in
Section 4 comes about, which is a natural question once that structure's
existence is established.

\subsection{Relation to prior work on cross-network comparison, alignment, and universality}

Our results suggest that quotienting out coordinate gauge provides a
principled basis for comparing independently trained networks. This
matters beyond this paper's specific question: whether the goal is to
ask about donor identity, to merge two models, or to compare mechanisms
discovered by separate training runs, raw parameter or neuron
correspondence risks conflating basis mismatch with genuine functional
difference. Any comparison that skips this step inherits whatever
arbitrary coordinate choice each training run happened to land on.

The question of whether independently trained networks converge to the
same internal structure predates the Neural Collapse literature, and has
been approached from several largely separate directions. For each, we
ask: what did they find, how does this paper's setting differ, and how
do the two results relate?

\textbf{Raw-coordinate comparison, before gauge correction.}
\citet{li2016convergent} were, to our knowledge, first to ask directly
whether independently trained networks learn the same representations,
using bipartite matching and spectral clustering to approximately align
neurons across independently trained CNNs. Their finding: units span
low-dimensional subspaces common across networks, while the specific
basis vectors are not. This is, in substance, the structure we report --
obtained years earlier, in a pre-Neural-Collapse setting, with discrete
neuron matching rather than a verified continuous affine map, and
without separating detectability from transplantability. Our
contribution relative to this line is not the qualitative shape of the
finding, which Li et al. anticipated, but a methodology that verifies
the alignment step exactly, measures and rules out its null-space
ambiguity as a confound, and operates in a regime where the shared
structure is independently characterized.

\textbf{Aggregate similarity indices.} \citet{kornblith2019similarity}
introduce CKA, which can reliably identify correspondences between
representations of differently initialized networks. CKA answers a
different question: it produces one scalar summarizing aggregate
similarity, not whether a specific residual can be attributed to a
specific donor. A high CKA score is consistent with either donor detail
being fully erased or a small but detectable residual surviving beneath
a large shared component; CKA alone cannot distinguish these. The
fingerprint test used here is a complementary, more targeted instrument
for exactly this distinction.

\textbf{Model stitching, which permits a trainable correction.}
\citet{bansal2021revisiting} connect the bottom layers of one network to
the top layers of another with a trainable layer at the seam, and report
that good networks can be stitched with little performance drop. We
frame this as a genuine contrast, not a contradiction -- though what
follows is our hypothesis for reconciling the two findings, not a claim
about what their trainable layer is proven to do: a trainable seam is,
in principle, free to absorb exactly the donor-specific detail this
paper tries to detect, since gradient descent on the stitched model
could use that layer to compensate for whatever the two representations
do not already share. We did not run a stitching experiment ourselves,
so this remains untested. Our own affine map is fit once, then held
fixed and verified exact; no further training is permitted at
evaluation time. Under the three-level distinction in Discussion
Section 3, stitching's question is closer to transplantability than to
detectability -- if this hypothesis is correct, the two findings answer
different levels of the same question rather than competing.

\textbf{Permutation symmetry and linear mode connectivity.} The
coordinate-gauge problem here is a continuous analogue of the discrete
permutation-invariance problem studied by \citet{entezari2022role} and
operationalized by \citet{ainsworth2023gitrebasin}, both asking whether
independently trained networks share a weight-space basin once
permutation symmetry is canonicalized away. Ainsworth et al. report the
first zero-barrier linear mode connectivity between independently
trained ResNets, but also a genuine counterexample showing it does not
hold universally. These results are about the geometry connecting
solutions, not about whether solutions differ in identifiable ways once
gauge is removed. This paper does not test mode connectivity; it takes
gauge removal as a prerequisite (Results 2) and asks what remains
detectable afterward.

\textbf{Neural Collapse theory and its transfer setting.} The
unconstrained features model of \citet{mixon2022neural} explains
theoretically why independently trained networks converge to the same
simplex-ETF geometry; we rely on this convergence empirically (Results
1, Discussion Section 4) without contributing new theory for why it
occurs. Separately, \citet{li2024understanding} study Neural Collapse in
transfer learning, finding collapse extent predicts downstream transfer
accuracy -- a different axis: they ask whether NC structure transfers
across \emph{tasks} for one trained network, while we ask whether
donor-specific structure survives alignment across independently trained
\emph{trajectories} on the same task.

\textbf{Universality in mechanistic interpretability.}
\citet{olah2020zoomin} speculate that features and circuits are
universal across independently trained models, citing convergent
learning as motivation. \citet{gurnee2024universal} test this directly
for neurons across five independently trained GPT-2 models, finding
only 1--5\% meet their universality threshold. This is compatible with
our result, but their comparison uses raw neuron activations without
any coordinate-gauge correction; part of their non-universality could
in principle reflect basis mismatch rather than genuine functional
difference. Our contribution here is narrow: after removing the
coordinate freedom raw neuron-level comparisons do not control for, and
controlling for a shared recipient-level component, donor-specific
structure remains reliably detectable -- suggesting some of the low
observed universality in neuron-level studies need not be attributed to
noise alone. This does not resolve universality in general; our setting
is a five-seed MLP-5 on MNIST, not a language model.

Whether the specific fingerprint construction used here (Methods 4.2) is
useful beyond the identification task it was built for is an open
question we have not tested. We raise this as a direction worth testing,
not as a demonstrated benefit; no merging, comparison, or
interpretability application was run in this paper.

\subsection{Limitations}

Five limitations bound the scope of what has been shown and should be
weighed directly against the central claim, not treated as footnotes to
it.

First, the training protocol is a paper-matched reconstruction of the
two-phase procedure of \citet{rupa2026neural}, not an exact replication: batch size and the
precise scheduler behavior at the Phase 1/2 transition could not be
verified against the original implementation (Methods 1).

Second, every result in this paper comes from a single architecture and
task -- an MLP-5 network on MNIST. Whether donor-specific fingerprints
remain identifiable, and whether the spectral-tail-pruning pattern of
Results 3 holds, in other architectures, other tasks, or other
collapse-inducing regimes is untested.

Third, five seeds were used throughout. This directly limits the
statistical resolution of the donor-identification test: the exact
permutation test's minimum attainable p-value is 1/120 regardless of the
true effect size, so the reported p=0.0083 should be read as the finest
statistical resolution this exact test can provide under a five-donor
design, not as an arbitrarily sharpenable figure.

Fourth, donor identification in this paper is based specifically on
per-class mean logit geometry (Methods 4.2). We did not compare this
construction against alternative fingerprint definitions, and we do not
claim it is the optimal or only viable choice; a different construction
could in principle yield different sensitivity to donor-specific
structure.

Fifth, and most consequential for the paper's central question: no
causal intervention was run. As discussed in Section 3, this paper
establishes detectability only. Whether a donor-specific fingerprint is
transplantable or causally persistent remains entirely open.

One additional observation is noted here without expansion: the
post-minimum rebound in NC2/NC3 described in Results 1 (Table~\ref{tab:table1b}) was
directionally consistent across seeds but varied substantially in onset
and magnitude, with one seed departing from the pattern shown by the
other four; we do not offer a mechanistic account for this difference,
and it does not bear on any result in Sections 1-5 above.

\subsection{Next causal test}

The fifth limitation above is not a gap to be noted and set aside; it
specifies the next experiment directly. Testing whether a detectable
fingerprint is transplantable and causally persistent requires: aligning
a donor's head into a recipient's coordinate frame using the construction
already verified in Methods 2; transplanting the aligned donor component
into the recipient network while explicitly controlling for recipient
norm and scale, so that any observed effect cannot be attributed to a
generic magnitude perturbation; resuming training on the recipient
network; and tracking, at intervals during continued training, whether
the donor's fingerprint persists, is progressively redirected toward the
recipient's own trajectory, or is erased. The same fingerprint definition
used in Stage 0D should be retained unless a change is explicitly
pre-registered, ensuring that the intervention is evaluated against the
same operational criterion that established detectability. Matched
perturbation and sham controls -- interventions of similar magnitude that
do not transfer any donor-specific structure -- are necessary to
distinguish a genuine transplant effect from a generic
perturbation-recovery response. The primary outcome measure for this
intervention should be pre-registered before it is run, consistent with
the confirmatory role such an experiment would play relative to the
exploratory findings reported here.

\subsection{Closing}

This paper set out to ask whether cross-trajectory variation remains
detectably donor-specific after coordinate gauge and recipient-level
structure are accounted for. Cross-trajectory variation can survive
removal of coordinate gauge as a detectable functional fingerprint.
Whether that fingerprint is portable under intervention remains the
decisive next question. The present work resolves the measurement
question; the intervention question remains.

\bibliographystyle{plainnat}
\bibliography{references}

\end{document}